\documentclass[10pt,twocolumn,letterpaper]{article}
\usepackage[svgnames,table]{xcolor}
\usepackage{iccv}
\usepackage{times}
\usepackage{epsfig}
\usepackage{graphicx}
\usepackage{color}
\usepackage{subfigure}
\usepackage[pagebackref=true,breaklinks=true,letterpaper=true,colorlinks,bookmarks=false]{hyperref}
\usepackage{amsmath,amsfonts,amssymb,mathrsfs}
\usepackage{makecell}

\usepackage{graphicx}
\usepackage[ruled]{algorithm2e}
\usepackage{amsmath,amssymb}
\usepackage{epstopdf}
\usepackage{tabularx}
\usepackage{eqnarray}
\usepackage{array,booktabs}
\usepackage{enumitem}
\usepackage[space]{grffile} 
\usepackage{soul} 
\usepackage{float}
\usepackage{tabularx,ragged2e} 
\usepackage{bbm}
\usepackage{pifont} 
\usepackage{cancel}
\usepackage{wrapfig}
\usepackage{color, colortbl}

\makeatletter

\renewcommand{\paragraph}{%
  \@startsection{paragraph}{4}%
  {\z@}{0.3ex \@plus 1ex \@minus .2ex}{-1em}
  {\normalfont\normalsize\bfseries}%
}
\makeatother

\definecolor{myGray}{rgb}{0.6,0.6,0.6}
\definecolor{myBlue}{rgb}{0.3,0.4,0.9}
\definecolor{myYellow}{rgb}{0.78,0.78,0.01}
\definecolor{myGreen}{rgb}{0.1,0.8,0.1}

\newcommand{\RR}{\mathbb{R}}

\def\onedot{\ifx\@let@token.\else.\null\fi\xspace}

\def\eg{\emph{e.g}\onedot} 
\def\ie{\emph{i.e}\onedot} 
 
\def\etc{\emph{etc}\onedot} 
\def\etal{\emph{et al}\onedot}

\definecolor{pcGray}{rgb}{0.5,0.5,0.5}

\definecolor{pccGray}{rgb}{0.3,0.3,0.3}

\newcommand{\fig}{Fig.~}
\newcommand{\eq}{Eq.\,}
\newcommand{\sect}{Section~}
\newcommand{\tab}{Table~}

\newcommand\closedots{\makebox[1em][c]{.\hfil.\hfil.}}

\usepackage{listings}
\usepackage{color}

\ifx \@etal \@empty \newcommand{\etal}{et al.} \fi
\ifx \@eg \@empty \newcommand{\eg}{e.g.,~} \fi
\ifx \@ie \@empty \newcommand{\ie}{i.e.,~} \fi
\ifx \@etc \@empty \newcommand{\etc}{etc.} \fi

\newcommand{\bh}{{\boldsymbol{h}}}

\newcommand{\bs}{{\boldsymbol{s}}}

\newcommand{\bw}{{\boldsymbol{w}}}

\iccvfinalcopy
\def\iccvPaperID{950}

\ificcvfinal\fi

\makeatletter
\let\@fnsymbol\@arabic
\makeatother

\begin{document}

\title{V-PROM: A Benchmark for Visual Reasoning Using\\Visual Progressive Matrices}

\author{Damien Teney*, Peng Wang*, Jiewei Cao,\\Lingqiao Liu, Chunhua Shen, Anton van den Hengel\\
Australian Institute for Machine Learning\\
The University of Adelaide\\
Adelaide, Australia\\
{\small*\textit{Authors with equal contribution in alphabetical order}}\\
{\tt\small \{firstname.lastname\}@adelaide.edu.au}
}

\maketitle

\begin{abstract}
One of the primary challenges faced by deep learning is the degree to which current methods exploit superficial statistics and dataset bias, rather than learning to generalise over the specific representations they have experienced.  This is a critical concern because generalisation enables robust reasoning over unseen data, whereas leveraging superficial statistics is fragile to even small changes in data distribution. To illuminate the issue and drive progress towards a solution, we propose a test that explicitly evaluates abstract reasoning over visual data. We introduce a large-scale benchmark of \emph{visual questions} that involve operations fundamental to many high-level vision tasks, such as comparisons of counts and logical operations on complex visual properties. The benchmark directly measures a method's ability to infer high-level relationships and to generalise them over image-based concepts. It includes multiple training/test splits that require controlled levels of generalization. We evaluate a range of deep learning architectures, and find that existing models, including those popular for vision-and-language tasks, are unable to solve seemingly-simple instances. Models using relational networks fare better but leave substantial room for improvement.
\end{abstract}


\section{Introduction}
\label{sec:intro}

\begin{figure}[t]
  \centering
  \includegraphics[width=0.9\linewidth]{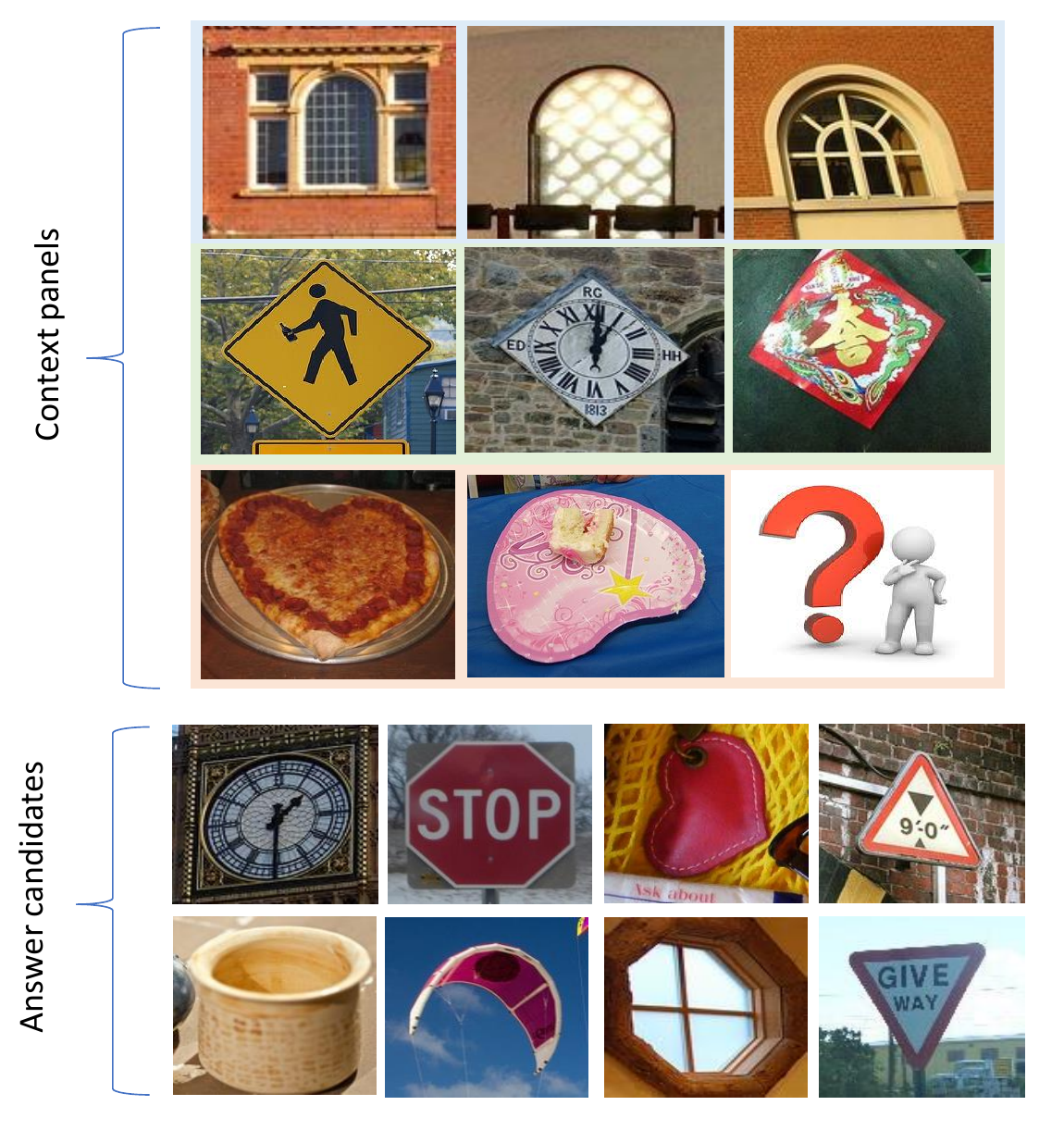}
  \vspace{-3pt}
  \caption{We propose a new task to evaluate a model's ability to perform abstract reasoning over complex visual stimuli. Each test instance is a matrix of $3\times3$ images, within which each row contains 3 images that exemplify the same relationship (in this case they have the same shape). The task is to identify the correct candidate for the missing $9^{\mbox{\tiny{th}}}$ image from a set of candidates. The correct answer above is the third candidate that represents a heart-shaped object.}
  \label{fig:teaser}
  \vspace{-13pt}
\end{figure}

Some of the most active research areas in computer vision are tackling increasingly complex tasks that require high-level reasoning. Some examples of this trend include visual question answering (VQA)~\cite{antol2015vqa}, image captioning~\cite{anderson2016guided}, referring expressions \cite{yu2017refexpr}, visual dialog~\cite{das2017dialog}, and vision-and-language navigation~\cite{mattersim}. While deep learning helped make significant progress, these tasks expose the limitations of the pattern recognition methods that have proved successful on classical vision tasks such as object recognition. A key indicator of the shortcomings of deep learning methods is their tendency to respond to specific features or biases in the dataset, rather than generalising to an approach that is applicable more broadly~\cite{vqacp,DevlinGGMZ15}. In response, we propose a benchmark to \emph{directly} measure a method's ability for high-level reasoning over real visual information, and in which we can control the level of generalisation required.

Progress on the complex tasks mentioned above is typically evaluated on standardized benchmarks~\cite{mattersim,antol2015vqa,chen2015microsoft,teney2016graphvqa}. Methods are evaluated with metrics on task-specific objectives, \eg predicting the correct answer in VQA, or producing a sentence matching the ground truth in image captioning. These tasks include a strong visual component, and they are naturally assumed to lie on the path to semantic scene understanding, the overarching goal of computer vision. Unfortunately, non-visual aspects of these tasks --~language in particular~-- act as major confounding factors. For example, in image captioning, the automated evaluation of generated language is itself an unsolved problem. In VQA, many questions are phrased such that their answers can be guessed without looking at the image.


We propose to take a step back with a task that directly evaluates abstract reasoning over realistic visual stimuli.  Our setting is inspired by Raven's Progressive Matrices (RPMs)~\cite{raven1938}, which are used in educational settings to measure human non-verbal visual reasoning abilities. Each instance of the task is a $3\times3$ matrix of images, where the last image is missing and is to be chosen from eight candidates. All rows of the completed matrix must represent a same relationship  (logical relationships, counts and comparisons, etc.) over a visual property of their three images~(\fig\ref{fig:teaser}). We use real photographs, such that the task requires strong visual capabilities, and we focus on visual, mostly non-semantic properties. This evaluation is thus designed to reflect the capabilities required by the complex tasks mentioned above, but in an abstract non-task-specific manner that might help guide general progress in the field.


Other recent efforts have proposed benchmarks for visual reasoning~\cite{pmlr-v80-barrett18a,suhr2017corpus} and our key difference is to focus on real images, which are of greater interest to the computer vision community than 2D shapes and line drawings. This is a critical difference, because abstract reasoning is otherwise much easier to achieve when applied to a closed set of easily identified symbols such as simple geometrical shapes. A major contribution of this paper is the construction of a suitable dataset with real images on large scale (over 300,000 instances).



Generalisation is a key issues that limits the robustness, and thus practicality of deep learning (see (\cite{groshev2018learning,fu2017learning,duan2017one,tobin2017domain} among many others). Current benchmarks that require visual reasoning, with few exceptions~\cite{vqacp,mattersim,tran2016rich}, use training and test splits that follow an identical distribution, which encourages methods to exploit dataset-specific biases (\eg class imbalance) and superficial correlations~\cite{jo2017measuring,szegedy2013intriguing}. This practice rewards methods that overfit to their training sets~\cite{vqacp} to the detriment of generalization capabilities. With these concerns in mind, our benchmark includes several evaluation settings that demand controlled levels of generalization (\sect~\ref{generalization}).



We have adapted and evaluated a range of deep learning models on our benchmark. Simple feed-forward networks achieve better than random results given enough depth, but recurrent neural networks and relational networks perform noticeably better. In the evaluation settings requiring strong generalization, \ie applying relationships to visual properties in combinations not seen during training, all tested models clearly struggle. In most cases, small improvements are observed by using additional supervision, both on the visual features (using a bottom-up attention network~\cite{anderson2017features} rather than a ResNet CNN~\cite{he2015resnet}), and on the type of relationship represented in the training examples. These results indicate the difficulty of the task while hinting at promising research directions.

Finally, \textbf{the proposed benchmark is not to be addressed as an end-goal, but should serve as a diagnostic test} of methods aiming at more complex tasks. In the spirit of the CLEVR dataset for VQA~\cite{johnson2016clevr} and the bAbI dataset for reading comprehension~\cite{weston2015babi}, our benchmark focuses on the fundamental operations common to multiple high-levels tasks. Crafting a solution specific to this benchmark is however not necessarily a path to actual solutions to these tasks. This guided the selection of general-purpose architectures evaluated in this paper.

\noindent
The contributions of this paper are summarized as follows.
\setlist{nolistsep,leftmargin=*}
\begin{enumerate}[noitemsep]
  \item We define a new task to evaluate a model's ability for abstract reasoning over complex visual stimuli. The task is designed to require reasoning similar to complex tasks in computer vision, while allowing evaluation free of task-specific confounding factors such as natural language and dataset biases.

  \item We describe a procedure to collect instances for this task at little cost, by mining images and annotations from the Visual Genome. We build a large-scale dataset of over 300,000 instances, over which we define multiple training and evaluation splits that require controlled amounts of generalization.

  \item We evaluate a range of popular deep learning architectures on the benchmark. We identify elements that prove beneficial (\eg relational reasoning and mid-level supervision), and we also show that all tested models struggle significantly when strong generalization is required.
\end{enumerate}
The dataset is publicly available on demand to encourage the development of models with improved capabilities for abstract reasoning over visual data.


\begin{figure*}[t!]
  \begin{center}
    \includegraphics[width=0.9\linewidth]{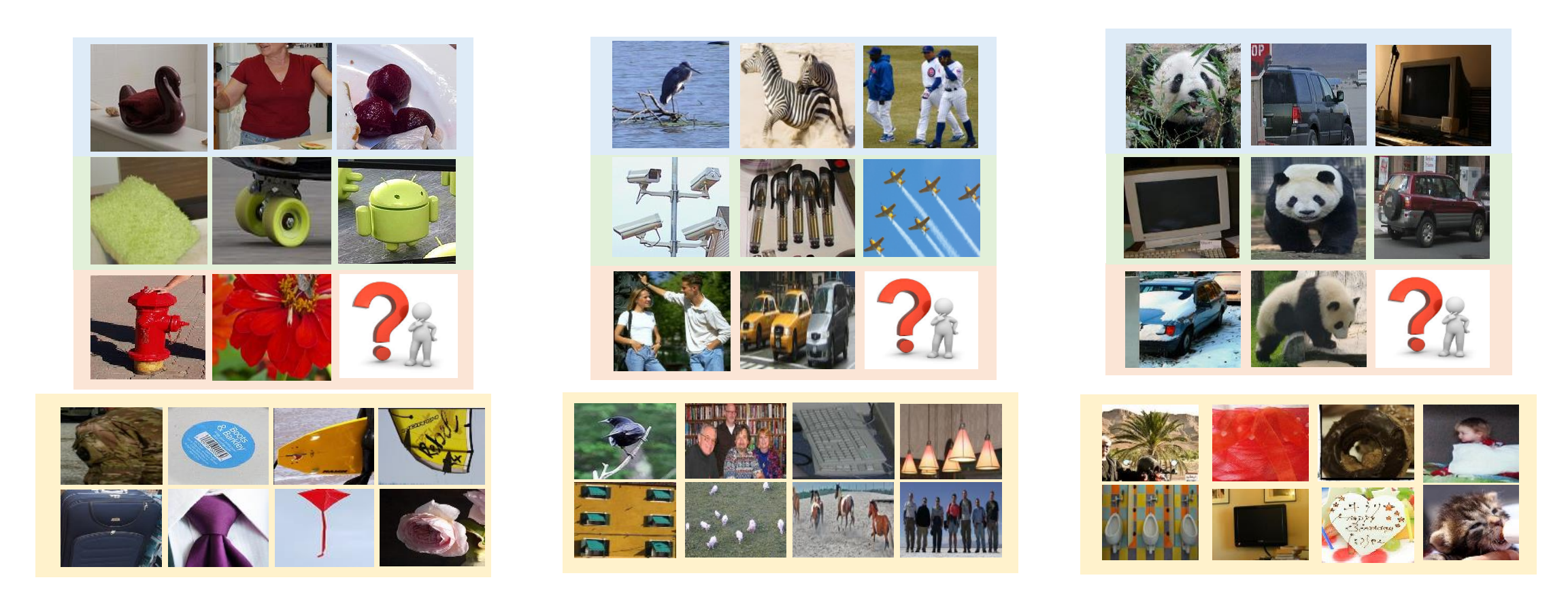}
  \end{center}
  \vspace{-12pt}
  \caption[]{Some challenging instances from our dataset. See the footnote\footnotemark for the answer key.}
  \vspace{-8pt}
  \label{fig:intro}
\end{figure*}

\section{Related work}
\label{relatedWork}

\paragraph{Evaluation of abstract visual reasoning} Evaluating reasoning has a long history in the field of AI, but is typically based on pre-defined or easily identifiable symbols. Recent works include the task set of Fleuret \etal~\cite{Fleureta2011ComparingMA}, in which they focus on the spatial arrangement of abstract elements in synthetic images. Their setting is reminiscent of the Bongard problems presented in~\cite{pr} and further popularized by Hofstadter~\cite{Hofstadter:1979:GEB:539932}. Stabinger \etal~\cite{10.1007/978-3-319-44781-0_45} tested whether state-of-the-art CNN architectures can compare visual properties of multiple abstract objects, \eg to determine whether two shapes are of the same size. Although this involves high-level reasoning, it is over coarse characteristics of line-drawings.

\begin{figure}[h!]
  \centering
  \includegraphics[width=1\linewidth]{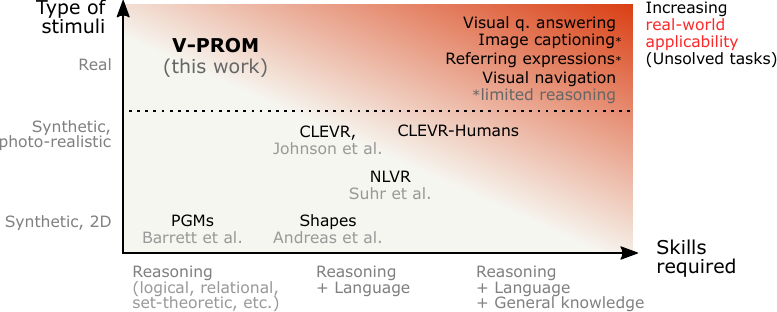}
  \vspace{0pt}
  \caption{Alternative tasks and datasets requiring visual reasoning. V-PROM fills an important gap between controlled, synthetic datasets (on which current methods are increasingly successful), and complex real-world tasks (which remain largely unsolved).}
  \label{fig:datasets}
  \vspace{-6pt}
\end{figure}

V-PROM is inspired by Raven's Progressive Matrices (RPMs)~\cite{raven1938}, a classic psychological test of a human's ability to interpret synthetic images. RPMs have been used previously to evaluate the reasoning abilities of neural networks~\cite{pmlr-v80-barrett18a,Hoshen2017IQON,Wang:2015:AGR:2832249.2832374}. In~\cite{Hoshen2017IQON}, the authors propose a CNN model to solve problems involving geometric operations such as rotations and reflections. Barrett \etal~\cite{pmlr-v80-barrett18a} evaluated existing deep learning models on a large-scale dataset of RPMs, with a procedure similar to one previously proposed by Wang \etal~\cite{Wang:2015:AGR:2832249.2832374}. The benchmark of Barrett \etal~\cite{pmlr-v80-barrett18a} is the most similar to our work. It uses synthetic images of simple 2D shapes, whereas ours uses much more complex images, at the cost of a less precise control of the visual stimuli. Recognizing the complementarity of the two settings, we purposefully model our evaluation setup after~\cite{pmlr-v80-barrett18a} such that future methods can be evaluated and compared across the two settings. Since the synthetic images in \cite{pmlr-v80-barrett18a} do not reflect the complexity of real-world data, progress on this benchmark may not readily translate to high-level vision tasks. Our work bridges the gap between these two extremes (Fig.\ref{fig:datasets}).



\paragraph{Evaluation of high-level tasks in computer vision} The interest in high-level tasks is growing, as exemplified by the advent of VQA~\cite{antol2015vqa}, referring expressions~\cite{yu2017refexpr}, and visual navigation~\cite{mattersim}, to name a few. Unbiased evaluations are notoriously difficult, and there is a growing trend toward evaluation on out-of-distribution data, \ie where the test set is drawn from a different distribution than the training set~\cite{vqacp,mattersim,teney2016zsvqa,tran2016rich}. In this spirit, our benchmark includes multiple training/test splits drawn from different distributions to evaluate generalization under controlled conditions. Moreover, our task focuses on abstract relationships applied to visual (\ie mostly non-semantic) properties, with the aim of minimizing the possibility of solving the task by exploiting non-visual factors. 

\paragraph{Models for abstract reasoning with neural networks} Various architectures have been proposed with the goal of moving beyond memorizing training examples, for example relation networks~\cite{NIPS2017_7082}, memory-augmented networks~\cite{DBLP:journals/corr/WestonCB14}, and neural Turing machines~\cite{DBLP:journals/corr/GravesWD14}. Recent works on meta learning~\cite{finn2017model,NIPS2016_6385} address the same fundamental problem by focusing on generalization from few examples (\ie few shot learning), and they have shown better generalization~\cite{oneShotimitation2018a}, including in VQA~\cite{Teney2017VisualQA}. Barrett \etal~\cite{pmlr-v80-barrett18a} applied relation networks (RNs) with success to their dataset of RPMs. We evaluate RNs on our benchmark with equally encouraging results, although there remains large room for improvement, in particular when strong generalization is required.

\vspace{4pt}

\footnotetext{Denoting the candidate answers as 1--8, left-to-right, first then second row, the correct ones are 7, 2, 6.} 

\section{A new task to evaluate visual reasoning}
\label{dataset}

Our task is inspired by the classical Raven's Progressive Matrices~\cite{raven1938} used in human IQ tests (see \fig\ref{fig:teaser}) . Each instance is a matrix of $3\times3$ images, where the missing final image must be identified from
among 8 candidates. The goal is to select an image such that all 3 rows represent a same relationship over some visual property (attribute, object category, or object count) of their 3 respective images. The definition of our task was guided by the following principles. First, it must require, but be not limited to, strong visual recognition ability. Second, it should measure a common set of capabilities required in high-level computer vision tasks. Third, it must be practical to construct a large-scale benchmark for this task, enabling an automatic and unambiguous evaluation. Finally, the task cannot be solvable through task-specific heuristics or relying on superficial statistics of the training examples. This points at a task that is compositional in nature and inherently requires strong generalization.

Our task can be seen as an extension to real images of recent benchmarks for reasoning on synthetic data~\cite{pmlr-v80-barrett18a,Hoshen2017IQON}. These works sacrifice visual realism for precise control over the contents of images which are limited to simple geometrical shapes. It is unclear whether reasoning under these conditions can transfer to realistic vision tasks. Our design is also intended to limit the extent to which semantic cues might be used to as ``shortcuts'' to avoid solving the task using the appropriate relationships. For example, a test to recognize the relation \emph{above} could rely on the higher likelihood of \emph{car above ground} than \emph{ground above car}, rather than its actual spatial meaning. Therefore, our task focuses on fundamental visual properties and relationships such as logical and counting operations over \emph{multiple} images (co-occurrence in a same photograph being likely biased).

The task requires identifying a plausible explanation for the provided triplets of images, \ie a relation that could have generated them. The incomplete triplet serves as a ``visual question'', and the explanation must be applied generatively to identify the missing image. It is unavoidable that more than one of the answer candidates constitute plausible completions. Indeed, a sufficiently-contrived explanation can justify any possible choice. The model has to identify the explanation with the strongest justification, which in practice tends to be the simplest one in the sense of Occam's razor. This is expected to be learned by the model from training examples.

\subsection{Construction of the V-PROM dataset}

We describe how to construct a large-scale dataset for our task semi-automatically. We it \emph{V-PROM} for \emph{Visual PROgressive Matrices}.

\begin{table}[t]
\small
\renewcommand{\tabcolsep}{0.25em}
\renewcommand{\arraystretch}{1.18}
  \centering
    \begin{tabularx}{\linewidth}{Xcccc}
    \hline
    &Object &Human & Object & Object\\
    & attributes & attributes & categories & counts\\
    \hline
    Nb. visual elements & 84 & 38    & 346     & 10 \\
    Nb. images & 36,750    & 12,249     & 82,905  & 11,730\\
    Nb. task instances & 45,000 & 45,000 & 45,000 & 100,000\footnotemark\\
    \hline
    \end{tabularx}
  \vspace{-1pt}
  \caption{Statistics of the V-PROM dataset.}
  \vspace{-13pt}
  \label{tab:stat}
\end{table}

\footnotetext{We generate more task instances with \textit{object counts} than with \textit{attributes} and \textit{categories} because counts are the only ones involved in the relationship \textit{progression}, in addition to the three others (\textit{and}, \textit{or}, \textit{union}).}

\label{instanceDescriptions}

\paragraph{Generating descriptions of task instances} Each instance is a matrix of $3\times{3}$ images that we call a visual reasoning matrix (VRM). Each image $I_i$ in the VRM depicts a visual element $a_i = \phi(I_i)$, where $a_i$ denotes an element depicted in the image with $a_i \in \mathcal{A}\cup\mathcal{O}\cup\mathcal{C}$, where $\mathcal{A}$, $\mathcal{O}$, $\mathcal{C}$, respectively denote sets of possible attributes, objects, and object counts. We denote with $v(I_i) \in \{A,O,C\}$ the type of visual element $a_i$ corresponds to. We also denote with $I_{i,j}$ the $j$-th image of the $i$-th row in a VRM. Each VRM represents one specific type $v$ of visual elements, and one specific type of relationship $r \in \{\textrm{And}, \textrm{Or}, \textrm{Union}, \textrm{Progression}\}$. We define them as follows.
\begin{itemize}
\item \textit{And}: $\phi(I_{i,3}) = \phi(I_{i,j}),~\forall j \in {1,2}$. The last image of each row has the same visual element as the other two. 
\item  \textit{Or}: $\phi(I_{i,3}) = \phi(I_{i,1})$ or $\phi(I_{i,3}) = \phi(I_{i,2})$. The last image in each row has the same visual element as the first or the second. 
\item \textit{Union}: $\{\phi(I_{1,j})~\forall~j\} = \{\phi(I_{2,j})~\forall~j\} = \{\phi(I_{3,j})~\forall~j\}$. All rows contain the same three visual elements, possibly in different orders.
\item \textit{Progression}: $v(I_{i,j})=C,~\forall i,j$; and $\phi(I_{i,t+1}) - \phi(I_{i,t}) = \phi(I_{j,t+1}) - \phi(I_{i,t})~\forall i,j, t \in{1,2}$. The numbers of objects in a row follow an arithmetic progression.
\end{itemize}
We randomly sample a visual element $v$ and relationship $r$ to generate the definition of a VRM. Seven additional incorrect answer candidates are obtained by sampling seven different visual elements of the same type as $v$. The following section describes how to obtain images that fulfills a definition $(v,r)$ of a VRM by mining annotations from the Visual Genome~(VG)~\cite{krishnavisualgenome}.

\paragraph{Mining images from the Visual Genome} To select suitable images, we impose five desired principles: \emph{richness}, \emph{purity}, \emph{image quality}, \emph{visual relatedness}, and \emph{independence}. Richness requires the diversity of visual elements, and of the images representing each visual element. Purity constrains the complexity of the image, as we want images that depict the visual element of interest fairly clearly. Visual relatedness guides us toward properties that have a clear visual depiction. As a counterexample, the attribute \emph{open} appears very differently when a door is open and a bottle is open. Such semantic attributes are not desirable for our task. Finally, independence excludes the objects that frequently co-occur with other objects (\eg ``sky'',``road'',``water'', \etc) and could lead to ambiguous VRMs.

We obtain images that fulfill the above principles using VG's region-level annotations of categories, attributes, and natural language description ( Table~\ref{tab:stat}). We first preselect categories and attributes with large numbers of instances to guarantee sufficient representations of each in our dataset.  We manually exclude unsuitable labels such as semantic attributes, and objects likely to cause ambiguity. We crop the annotated regions to obtain \emph{pure} images. We discard those smaller than 100~px in either dimension. The annotations of \emph{object counts} are extracted from numbers 1--10 appearing in natural language descriptions (\eg ``five bowls of oatmeal''), manually excluding those unrelated to counts (\eg ``five o'clock'' or ``a 10 years old boy'').

\subsection{Data splits to measure generalization}
\label{generalization}

In order to evaluate a method's capabilities for generalization, we define several training/evaluation splits that require different levels of generalization. Training and evaluating a method in each of these settings will provide an overall picture of its capabilities beyond the basic fitting of training examples. To define these different settings, we follow the nomenclature proposed by Barrett \etal~\cite{pmlr-v80-barrett18a}.

\setlist{nolistsep,leftmargin=*}
\begin{enumerate}[noitemsep]
  \item \textbf{Neutral} -- The training and test sets are both sampled from the whole set of relationships and visual elements. Training to testing ratio is $2:1$.

  \item \textbf{Interpolation / extrapolation} -- These two splits evaluate generalization for counting. In the interpolation split, odd counts (1,3,5,7,9) are used for training and even counts (2,4,6,8,10) are used for testing. In the extrapolation split, the first five counts (1--5) are used for training and the remaining (6--10) are used for testing.

  \item \textbf{Held-out attributes} -- The object attributes are divided into 7 super-attributes\footnote{The attributes within each super-attribute are mutually exclusive.}: color, material, scene, plant condition, action, shape, texture. The human attributes are divided into 6 super-attributes: age, hair style, clothing style, gender, action, clothing color. The super-attributes \emph{shape, texture, action} are held-out for testing only.

  \item \textbf{Held-out objects} -- A subset of object categories ($1/3$) are held-out for testing only.

  \item \textbf{Held-out pairs of relationships/attributes} -- A subset of relationship/super-attribute combinations are held-out for testing only. Three combinations are held-out for both object attributes and human attributes. The held-out super-attributes vary with each type of relationship.

  \item \textbf{Held-out pairs of relationships/objects} -- For each type of relationship, $1/3$ of objects are held-out. The held-out objects are different for each relationship. 
\end{enumerate}

\noindent
We report a model's performance with the accuracy, \ie the fraction of test instances for which the predicted answer (among the eight candidates) is correct. Random guessing gives an accuracy of $12.5\%$.

\subsection{Task complexity and human evaluation}
Solving an instance of our task requires to recognize the visual elements depicted in all images, and to identify the relation that applies to triplets of images. This basically amounts to inferring the abstract description (\sect\ref{instanceDescriptions}) $\mathcal{S}=\{[r,v]:r\in\mathcal{R},v\in\mathcal{V}\}$ of the instance. Our dataset contains 4 types of relations, applied over 478 types of visual elements (\tab~\ref{tab:stat}), giving in the order of 2,000 different combinations.

We performed a human study to assess the difficulty of our benchmark. We presented human subjects with a random selection of task instances, sampled evenly across the four types of relations. The testees can skip an instance if they find it too difficult or ambiguous. The accuracy was of $77.8\%$ with a \emph{skip} rate of $4.5\%$. This accuracy is not an upper bound for the task however. The two main reasons for non-perfect human performance are (1)~counting errors with $>$5 objects, cluttered background, or scale variations and (2)~a tendency to use prior knowledge and favor higher-level (semantic) concepts/attributes than those used to generate the dataset.


\section{Models and experimental setup}

\abovedisplayskip=6pt
\belowdisplayskip=6pt

\begin{figure}[t]
  \centering
  \includegraphics[width=0.99\linewidth]{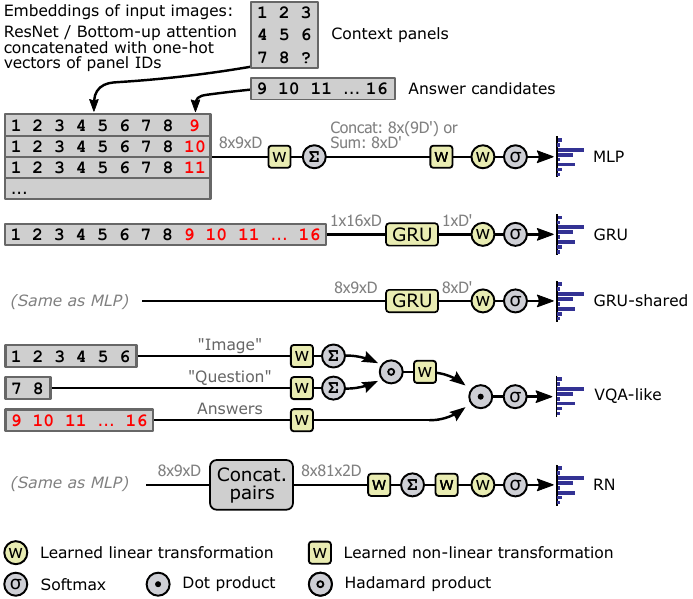}
  \vspace{0pt}
  \caption{Overview of the models evaluated in our experiments. These are based on popular deep learning architectures.}
  \label{fig:models}
  \vspace{-10pt}
\end{figure}

We evaluated a range of models on our benchmark. These models are based on popular deep learning architectures that have proven successful on various task-specific benchmarks. The models are summarized in \fig\ref{fig:models}.

\subsection{Input data}
For each instance of our task, the input data consists of 8 context panels and 8 candidate answers. These 16 RGB images are passed through a pretrained CNN to extract visual features. Our experiments compare features from a ResNet101~\cite{he2015resnet} and from a Bottom-Up Attention Network\footnote{The network of \etal\cite{anderson2017features} was pretrained with annotations from the Visual Genome. Our dataset only uses cropped images from VG, and we use layer activations rather than explicit class predictions, but the possible overlap in the label space used to pretrain \cite{anderson2017features} and to generate our benchmark must be kept in mind.}~\cite{anderson2017features}, which is popular for image captioning and VQA~\cite{teney2017challenge}. The feature maps from either of these CNNs are average-pooled, and the resulting vector is L2-normalized. The vector of each of the 16 images is concatenated with a one-hot representation of an index: the 8 context panels are assigned indices 1--8 and the candidate answers 9--16. The resulting vectors are referred to as $x_1, x_2, ... x_{16} \in \RR^{2048+16}$.

The vectors $x_i$ serve as input to the models described below, which are trained with supervision to predict a score for each of the 8 candidate answers, \ie $\hat{\bs} \in \RR^8$. Each model is trained with a softmax cross-entropy loss over $\hat{\bs}$, standard backpropagation and SGD, using AdaDelta~\cite{zeiler2012adadelta} as the optimizer. Suitable hyperparameters for each model were coarsely selected by grid search (details in supplementary material). We held out 8,000 instances from the training set to serve as a validation set, to select the hyperparameters and to monitor for convergence and early-stopping. Unless noted, the non-linear transformations within the networks below refer to a linear layer followed by a ReLU.

\subsection{MLP}
Our simplest model is a multilayer perceptron (see \fig\ref{fig:models}). The features of every image are passed through a non-linear transformation $f_1(\cdot)$. The model is then applied so as to share the parameters used to score each candidate answer. The features of each candidate answer ($x_i$ for $i$=$9,\closedots,16$) are concatenated with the context panels ($x_1,\closedots,x_8$). The features are then passed through another non-linear transformation  $f_2(\cdot)$, and a final linear transformation $\bw$ to produce a scalar score for each candidate answer. That is, $\forall~i = 1\closedots8$:
\begin{align}
  \hat{s}_i ~=~ \bw ~ f_2\big( [ f_1(x_1) ; f_1(x_2) ; \closedots ; f_1(x_8) ; f_1(x_{8+i}) ] \big)
  \label{eq:mlpAvg}
\end{align}
where the semicolumn represents the concatenation of vectors. A variant of this model replaces the concatenation with a sum-pooling over the nine panels. This reduces the number of parameters by sharing the weights within $f_2$ across the panels. This gives
\begin{align}
  \hat{s}_i ~=~ \bw ~ f_2\big( \; \Sigma_{i=1,2,\closedots,x_8,8+i} ~f_1(x_i) \; \big) ~~.
  \label{eq:mlpSum}
\end{align}
We will refer to these two models as \textit{MLP-cat-k} and \textit{MLP-sum-k}, in which $f_1$ and $f_2$ are both implemented with $k/2$ linear layers, all followed by a ReLU.

\subsection{GRU}
We consider two variants of a recurrent neural network, implemented with a gated recurrent unit (GRU~\cite{cho2014learning}). The first naive version takes each of the feature vectors $x_1$ to $x_16$ over 16 time steps. The final hidden state of the GRU is then passed through a linear transformation $\bw$ to map it to a vector of 8 scores $\hat{\bs} \in \RR^8$.
\begin{align}
  \hat{\bs} ~=~ \bw ~ \textrm{GRU}\big(x_1, x_2, \closedots, x_8, x_9, x_{10}, \closedots, x_{16} \big) ~.
  \label{eq:gru}
\end{align}
The second version shares the parameters of the model over the 8 candidate answers. The GRU takes, in parallel, 8 sequences, each consisting of the context panels with one of the 8 candidate answers. The final state of each GRU is then mapped to a single score for the corresponding candidate answer. That is, $\forall~i = 1\closedots8$:
\begin{align}
  \hat{s}_i ~=~ \bw ~ \textrm{GRU}\big(x_1, x_2, \closedots, x_8, x_{8+i} \big) ~.
  \label{eq:gruShared}
\end{align}

\subsection{VQA-like architecture}
We consider an architecture that mimics a state-of-the-art model in VQA~\cite{teney2017challenge} based on a ``joint embedding'' approach~\cite{wu2017survey,teney2017spm}. In our case, the context panels $x_1, \closedots, x_6$ serve as the input ``image'', and the panels $x_7,x_8$ serve as the ``question''. They are passed through non-linear transformations, then combined with an elementwise product into a joint embedding $\bh$. The score for each answer is obtained as the dot product between $\bh$ and the embedding of each candidate answer (see \fig\ref{fig:models}). Formally, we have
\begin{align}
  \bh& ~=~ \Sigma_{i=1\closedots6} ~ f_1(x_i) ~~\circ~~ \Sigma_{i=7,8} ~f_2(x_i)\\
  \hat{s}_i& ~=~ \bh . f_3(x_{8+i})~.
  \label{eq:vqaLike}
\end{align}
where $f_1$, $f_2$ and $f_3$ are non-linear transformations, and $\circ$ represents the Hadamard product.

\subsection{Relation networks}
We finally evaluate a relation network (RN). RNs were specifically proposed to model relationships between visual elements, such as in VQA when questions refer to multiple parts of the image~\cite{atzmon2016relational}. Our model is applied, again, such that its parameters are shared across answer candidates. The basic idea of an RN is to consider all pairwise combinations of input elements ($9^2$ in our case), pass them through a non-linear transformation, sum-pool over these $9^2$ representations, then pass the pooled representation through another non-linear transformation. Formally, we have, $\forall~i = 1\closedots8$:
\begin{align}
  \bh_i &~=~ \Sigma_{(i,j) \in \{1,2,\closedots,8,8+i\}} f_1([ x_i ; x_j ])\\
  \hat{s}_i &~=~ \bw ~ f_2( \bh_i )
  \label{eq:rn}
\end{align}
where $f_1$ and $f_2$ are non-linear transformations, and $\bw$ a linear transformation.

\subsection{Auxiliary objective}
We experimented with an auxiliary objective that encourages the network to predict the type of the relationship involved in the given matrix. This objective is trained with a softmax cross-entropy and the ground truth type of relationship in the training example. This value is a index among the seven possible relations, \ie \textit{and}, \textit{or}, \textit{progression}, \textit{attribute}, \textit{object}, \textit{union}, and \textit{counting} (see \sect\ref{dataset}). This prediction is made from a linear projection of the final activations of the network in \eq\ref{eq:rn}, that is:
\begin{align}
  \hat{s}_i &~=~ \bw' ~ f_2( \bh )
  \label{eq:aux}
\end{align}
where $\bw'$ is an additional learned linear transformation. At test time, this prediction is not used, and the auxiliary objective serves only to provide an inductive bias during the training of the network such that its internal representation captures the type of relationship (which should then help the model to generalize). Note that we also experimented with an auxiliary objective for predicting labels such as object class and visual attributes, but this did not prove beneficial.

\begin{figure}[t!]
  \centering
  \includegraphics[width=0.99\linewidth]{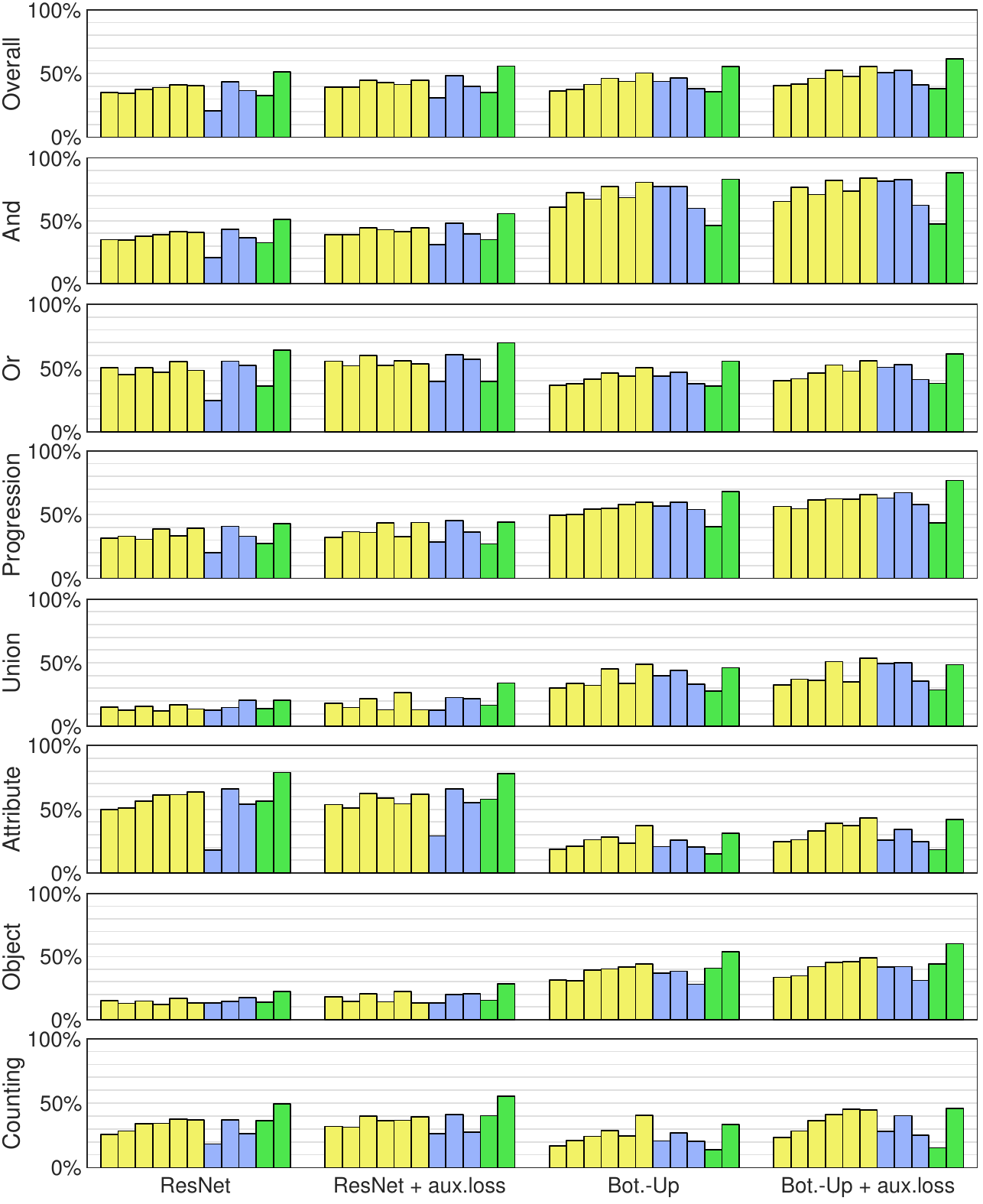}
  \vspace{2pt}
  \caption{Accuracy of all modes in the neutral setting, broken down by question type. The types \textit{and/or/progression/union} reflect the type of relationship across the nine images, while \textit{attribute/object/counting} correspond to the type of visual properties to which the relationship applies. Each group of bars corresponds to the methods \textit{\textcolor{myYellow}{MLP-cat-2}, \textcolor{myYellow}{MLP-cat-4}, \textcolor{myYellow}{MLP-cat-6}, \textcolor{myYellow}{MLP-sum-2}, \textcolor{myYellow}{MLP-sum-4}, \textcolor{myYellow}{MLP-sum-6}, \textcolor{myBlue}{GRU}, \textcolor{myBlue}{GRU-shared}, \textcolor{myBlue}{VQA-like}, \textcolor{myGreen}{RN without panel IDs}, and \textcolor{myGreen}{RN}}. See supplementary material for numbers.}
  \label{fig-bar1}
  \vspace{-8pt}
\end{figure}

\begin{figure}[t!]
  \centering
  \includegraphics[width=0.99\linewidth]{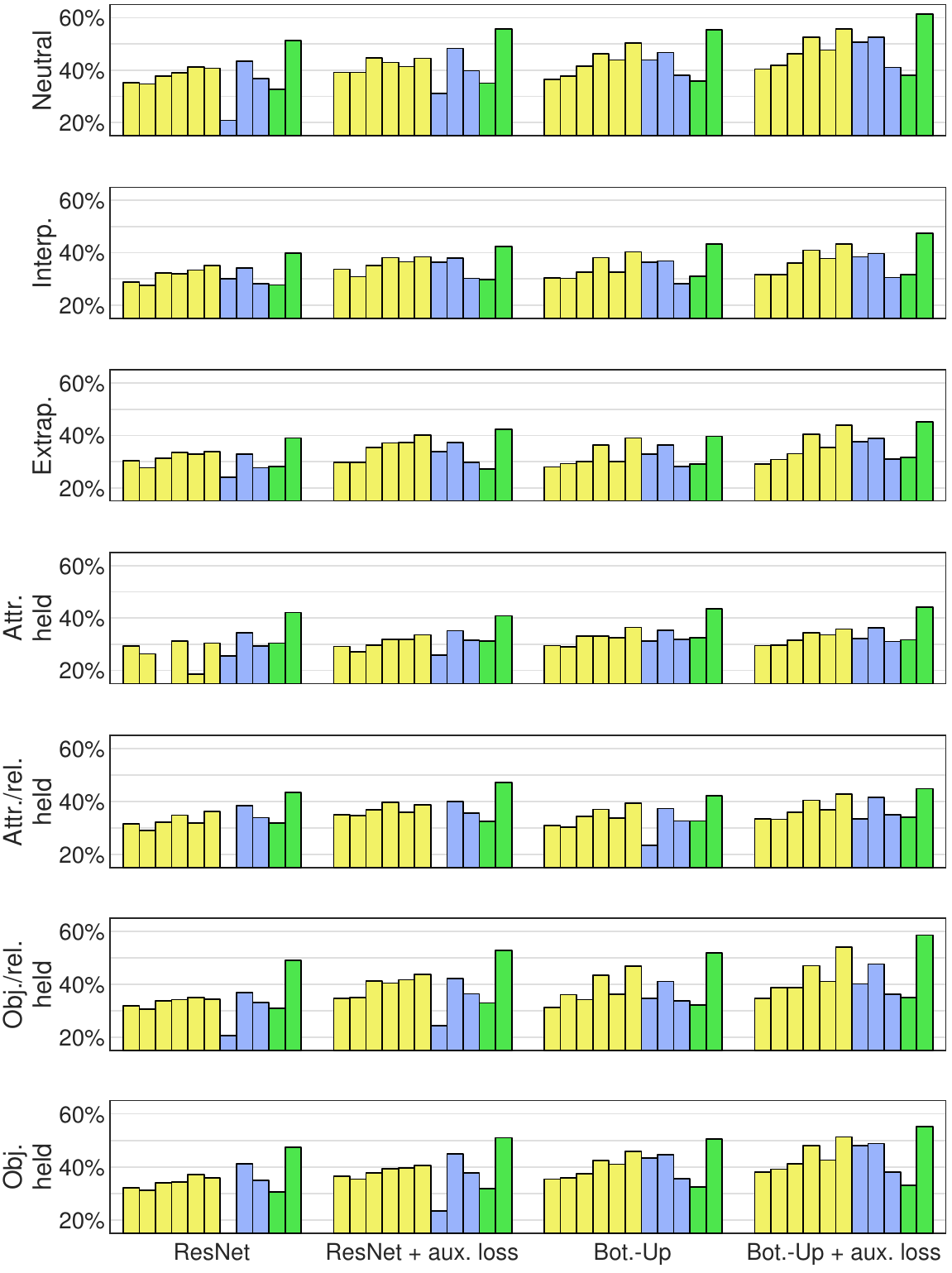}
  \vspace{2	pt}
  \caption{Accuracy of all models trained/evaluated on splits requiring varying levels of generalization. The relative performance of the models is generally consistent, but all models perform significantly worse than in the \textit{neutral} setting, indicating poor generalization of most models and overfitting to their training examples. Each group of bars corresponds to the same methods as in \fig\ref{fig-bar1}.}
  \label{fig-bar2}
  \vspace{-8pt}
\end{figure}

\section{Experiments}

We conducted numerous experiments to establish reference baselines and to shed light on the capabilities of popular architectures. As a sanity check, we trained our best model with randomly-shuffled context panels. This verified that the task could not be solved by exploiting superficial regularities of the data. All models trained in this way perform around the ``chance'' level of 12.5\%.

\begin{table}[t]
\small
\renewcommand{\tabcolsep}{0.15em}
\renewcommand{\arraystretch}{1.18}
\centering
\begin{tabularx}{\linewidth}{Xc cccc}
\Xhline{1\arrayrulewidth}
                                          &   ResNet & ResNet & B.-up & B.-up\\
                                          &   ~ & +aux.loss & ~& +aux.loss\\
\Xhline{1\arrayrulewidth}
Human evaluation& \multicolumn{4}{c}{77.8}\\
RN with	 shuffled inputs& 12.5 & 12.5 & 12.5 & 12.5\\ 
\Xhline{1\arrayrulewidth}
MLP-sum-6 layers& 40.7 & 44.5 & 50.4 & 55.7\\
GRU-shared& 43.4 & 48.2 & 46.7 & 52.7\\
VQA-like& 36.7 & 39.7 & 37.9 & 41.0\\
Relational network (RN)& \textbf{51.2} & \textbf{55.8} & \textbf{55.4} & \textbf{61.3}\\
\Xhline{1\arrayrulewidth}
\end{tabularx}
\normalsize
\normalsize
\caption{Summary of the best models in the neutral setting, on all question types (\fig\ref{fig-bar1}, first row). Additional results in supp.mat.}
\vspace{-9pt}
\label{tableNeutralSummary}
\end{table}

\subsection{Neutral training/test splits}
\label{expNeutral}
We first examine all models on the neutral training/test splits (\fig\ref{fig-bar1} and \tab\ref{tableNeutralSummary}). In this setting, training and test data are drawn from the same distribution, and supervised models are expected to perform well, given sufficient capacity and training examples. We observe that a \textbf{simple MLP} can indeed fit the data relatively well if it has enough layers, but a network with only 2 non-linear layers performs quite badly. The two models based on a \textbf{GRU} have very different performance. The \textit{GRU-shared} model performs best. It shares its parameters over the candidate answers (processed in parallel rather than across the recurrent steps). This result was not obviously predictable, since this model does not get to consider all candidate answers in relation with each other. The alternate model (\textit{GRU}) receives every candidate answer in succession. It could therefore perform additional reasoning steps over the candidates, but this does not seem to be the case in practice. The \textbf{VQA-like} model obtains a performance comparable to a deep MLP, but it proved more difficult to train than an MLP. In some of our experiments, the optimization this model was slow or simply failed to converge. We found it best to use, as non-linear transformations, ``gated tanh'' layers as in~\cite{teney2017challenge}. Overall, we obtained the best performance with a \textbf{relation network} (RN) model. While this is basically an MLP on top of pairwise combinations of features, these combinations prove much more informative than the individual features. We experimented with an RN without the one-hot representations of panel IDs concatenated with the input (``\textit{RN without panel IDs}''), and this version performed very poorly. It is worth noting that RNs come at the cost of processing $N^2$ feature vectors rather than $N$ (with $N$=9 in our case). The number of parameters is the same, since they are shared across the $N^2$ combinations, but the computation time increases.

We break down performance along two axes in \fig\ref{fig-bar1}. The following two groups of question types are mutually exclusive: and/or/progression/union, and attribute/object/counting. The former reflects the type of relationship across the nine images of a test instance, while the latter corresponds to the type of visual properties to which the relationship applies. We observe that some types are much more easily solved than others. Instances involving object identity are easier than those involving attributes and counts, presumable because the image features are obtained with a CNN pretrained for object classification. The bottom-up image features performs remarkably well, most likely because the set of labels used for pretraining was richer than the ImageNet labels used to train the ResNet. The instances that require counting are particularly difficult; this corroborates the struggle of vision systems with counting, already reported in multiple existing works, \eg in~\cite{kafle2017analysis}.

\subsection{Splits requiring generalization}
\label{expGeneralization}
We now look at the performance with respect to splits that specifically require generalization (\fig\ref{fig-bar2}). As expected, accuracy drops significantly as the need for generalization increases. This confirms our hypothesis that naive end-to-end training cannot guarantee generalization beyond training examples, and that this is easily masked when the test and training data come from the same distribution (as in the neutral split). This drop is particularly visible with the simple MLP and GRU models. The RN model suffers a smaller drop in performance in some of the generalization settings. This indicates that learning over combinations of features provides a useful inductive bias for our task.

\paragraph{Image features from bottom-up attention}
We tested all models with features from a ResNet, as well as features from the ``bottom-up attention'' model of Anderson \etal~\cite{anderson2017features}. These improve the performance of all tested models over ResNet features, in the neutral and all generalization splits. The bottom-up attention model is pretrained with a richer set of annotations than the ImageNet labels used to pretrain the ResNet. This likely provides features that better capture fine visual properties of the input images. Note that the visual features used by our models do not contain explicit predictions of such labels and visual properties, as they are vectors of continuous values. We experimented with alternative schemes (not reported in the plots), including an auxiliary loss within our models for predicting visual attributes, but these did not prove helpful.

\paragraph{Auxiliary prediction of relationship type}
We experimented with success with an auxiliary loss on the prediction of the type of relationship in the given instance. This is provided during training as a label among seven. All models trained with this additional loss gained in accuracy in the neutral and most generalization settings. The relative importance of the main and auxiliary losses did not seem critical, and all reported experiments use an equal weight on both.

Overall, the performance of our best models remains well below that of human performance leaving substantial room for improvement. This dataset should be a valuable tool to evaluate future approaches to visual reasoning.

\section{Conclusions}

We have introduced a new benchmark to measure a method's ability to carry out abstract reasoning over complex visual data. The task addresses a central issue in deep learning, being the degree to which methods learn to reason over their inputs. This issue is critical because reasoning can generalise to new classes of data, whereas memorising incidental relationships between signal and label does not. This issue lies at the core of many of the current challenges in deep learning, including zero-shot learning, domain adaptation, and generalisation, more broadly.


Our benchmark serves to evaluate capabilities similar to some of those required in high-level tasks in computer vision, without task-specific confounding factors such as natural language or dataset biases. Moreover, the benchmark includes multiple evaluation settings that demand controllable levels of generalization. Our experiments with popular deep learning models demonstrate that they struggle when strong generalization is required, in particular for applying known relationships to combinations of visual properties not seen during training. We identified a number of promising directions for future research, and we hope that this setting will encourage the development of models with improved capabilities for abstract reasoning over visual data.


{\small\bibliographystyle{ieee}\bibliography{Bibliography}}
\clearpage

\appendix
\section*{Supplementary material}
\vspace{4pt}




\section{Dataset details}
Table~\ref{tab:stat_supp} shows the number of training/test instances in the different data splits. 

\vspace{-3pt}
\begin{table}[h]
\label{tab:num_ins}
\small
\renewcommand{\tabcolsep}{0.53em}
\renewcommand{\arraystretch}{1.18}
  \centering
    \begin{tabularx}{\linewidth}{Xcc}
    \hline
    &Training &Test\\
    \hline
    Neutral & 103,323 & 51,677 \\
    Interpolation & 109,991 & 65,009 \\
    Extrapolation & 109,991 & 65,009 \\
    Att.held & 103,329 & 51,617 \\
    Att.rel.held &73,329 & 51,671 \\
    Obj.held &103,326 & 51,674\\
    Obj.rel.held &88,326 & 51,674\\
    
    \hline
    \end{tabularx}
  \vspace{-1pt}
  \caption{Number of training/test instances in each data split.}
  \vspace{-13pt}
  \label{tab:stat_supp}
\end{table}

\section{Implementation details}


The \textbf{image features} were obtained with the ResNet-101 CNN~\cite{he2015resnet} implemented in MXNet~\cite{mxnet} and pretrained on ImageNet~\cite{imagenet}, and with the Bottom-Up Attention network of Anderson \etal~\cite{anderson2017features}. The latter uses an R-CNN framework itself based on a ResNet-101. We resize each image of our dataset such that its shorter side is of 256 pixels, and preprocess it with color normalization. We then crop out the central $224 \times 224$ patch from the resulting image and feed it to the network. Feature maps from the last convolutional layer are pre-extracted in this way for every image. These feature maps are pooled (averaged) over image locations (with the ResNet) or over region proposals (for the Bottom-Up Attention network). The resulting vector is of dimension 2048, and is \textbf{normalized to unit $L_2$ length}. The normalization of the image features is crucial to obtain reasonable performance. This has previously been reported for other tasks like VQA.

All models are trained with a \textbf{batch size} of 128, and a size of all \textbf{hidden layers} of 128. These values were selected by grid search and performed consistently well across models. All models use the one-hot labels of the input panels, except the model referred to as ``RN without panel IDs''. All models are optimized using \textbf{AdaDelta}~\cite{zeiler2012adadelta}.

Models using an \textbf{auxiliary loss} use the same weight for the two losses. We experimented with different relative weights, and it did not affect the results significantly in either direction. All \textbf{non-linear layers} are implemented with affine weights followed by a ReLU, except in the VQA-like model, which proved easier to optimize with ``gated tanh'' layers, as in the VQA model of Teney \etal~\cite{teney2017challenge}.

Let us also mention that we experimented with a VQA-like network that includes a \textbf{top-down attention} mechanism as in~\cite{teney2017challenge}. This performed slightly worse than the simple model, and it is not included in our results.

\section{Additional results}

We provide in Tables \ref{tableNeutral} and \ref{tableGeneralization} all numbers corresponding to the bar plots presented in the paper.

\begin{figure}[h!]
  \centering
  \includegraphics[width=0.80\linewidth]{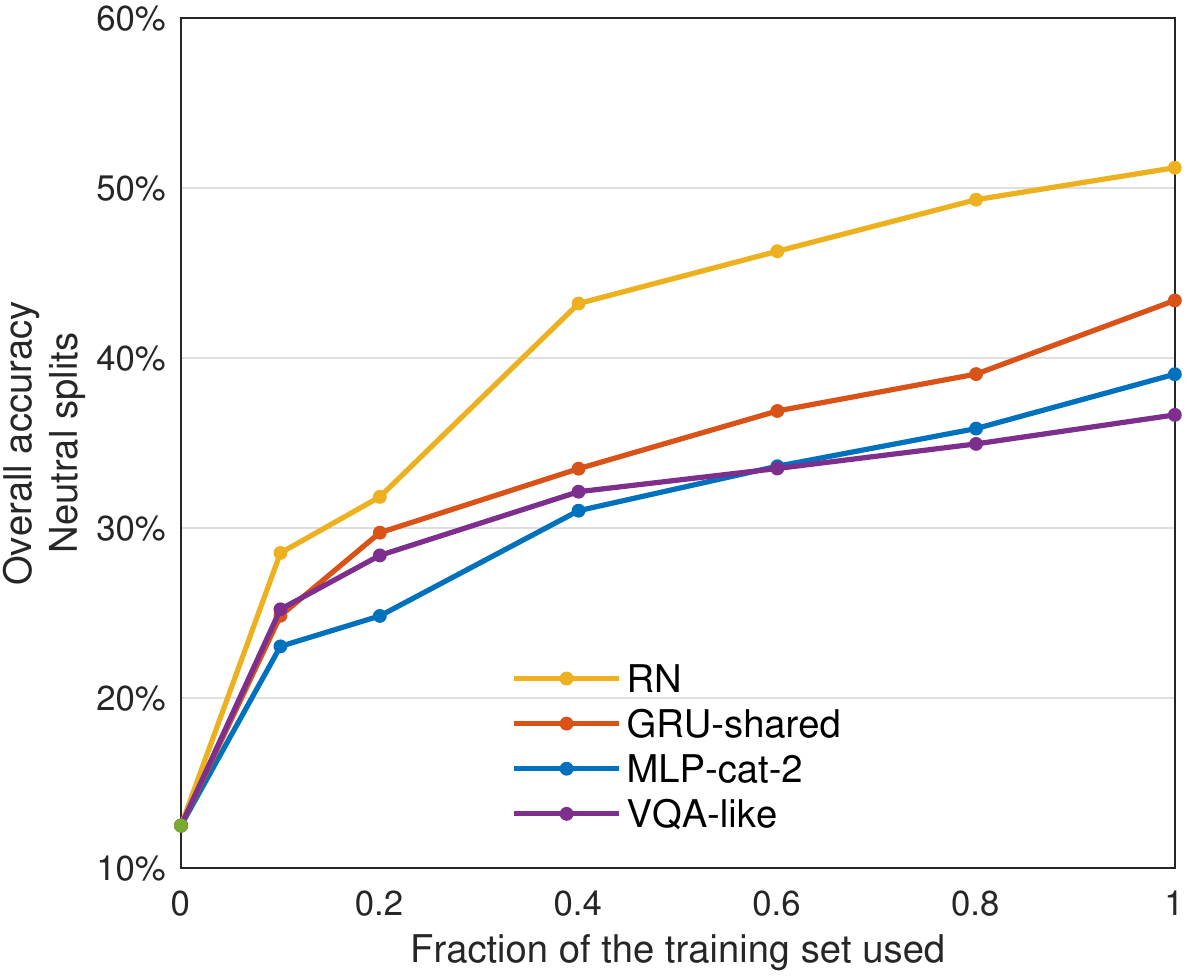}
  \caption{Accuracy of various models trained on reduced amounts of training data.}
  \label{fig-reducedTr}
  \vspace{-8pt}
\end{figure}

We performed additional experiments to compare the sample efficiency of the different models. We trained our best models with only a fraction of the training set. We included 4 of our best models, using simple ResNet features and without the auxiliary loss. The results are presented in \fig\ref{fig-reducedTr}. The sample efficiency is fairly consistent across models. The performance grows almost linearly with the amounts of training data, which indicates the explicit reliance of these models on the training examples (\ie their weak ability to generalize). These results can also serve as supplementary baselines to investigate low-data regimes 

\begin{table*}[t]
\small
\renewcommand{\tabcolsep}{0.15em}
\renewcommand{\arraystretch}{1.18}
\centering
\begin{tabularx}{\linewidth}{Xc c cccccccc}
\Xhline{1\arrayrulewidth}
                                          &   Overall & ~~ & \multicolumn{7}{c}{Accuracy per question type} \vspace{-0pt}\\ \cline{4-5} \cline{4-7} \cline{9-11}
                                          &   accuracy & ~~~ & \textit{~and~~} & \textit{~~~or~~} & \textit{progression} & \textit{~union~} & ~~~& \textit{attribute} & \textit{~object~} & \textit{~counting} \\
\Xhline{1\arrayrulewidth}
Human evaluation & 77.8 & ~ & -- & -- & -- & -- & ~ & -- & -- & --\\
Lower bound: RN (last row) with shuffled inputs & 12.5 & ~ & 12.5 & 12.5 & 12.5 & 12.5 & ~ & 12.5 & 12.5 & 12.5\\ 
\Xhline{1\arrayrulewidth}
MLP-sum-2, ResNet	 & 35.1 & ~ & 35.1 & 50.2 & 31.4 & 15.1 & ~ & 49.7 & 15.1 & 25.7\\
MLP-sum-2, ResNet + aux. loss	 & 39.1 & ~ & 39.1 & 55.5 & 32.1 & 18.0 & ~ & 53.7 & 17.9 & 31.9\\
MLP-sum-2, Bot.-Up	 & 36.4 & ~ & 60.8 & 36.4 & 49.4 & 30.1 & ~ & 18.7 & 31.6 & 16.8\\
MLP-sum-2, Bot.-Up + aux. loss	 & 40.4 & ~ & 65.5 & 40.4 & 56.5 & 32.8 & ~ & 24.4 & 33.6 & 23.4\\
\Xhline{1\arrayrulewidth}
MLP-cat-2, ResNet	 & 34.7 & ~ & 34.7 & 45.0 & 33.0 & 12.8 & ~ & 51.2 & 12.9 & 28.3\\
MLP-cat-2, ResNet + aux. loss	 & 39.2 & ~ & 39.2 & 51.8 & 36.7 & 14.9 & ~ & 51.0 & 14.4 & 31.4\\
MLP-cat-2, Bot.-Up	 & 37.6 & ~ & 72.3 & 37.6 & 50.0 & 33.8 & ~ & 20.8 & 30.7 & 21.0\\
MLP-cat-2, Bot.-Up + aux. loss	 & 41.7 & ~ & 76.7 & 41.7 & 54.6 & 37.1 & ~ & 26.1 & 34.8 & 28.4\\
\Xhline{1\arrayrulewidth}
MLP-sum-4, ResNet	 & 37.6 & ~ & 37.6 & 50.2 & 30.6 & 15.6 & ~ & 56.2 & 14.7 & 34.2\\
MLP-sum-4, ResNet + aux. loss	 & 44.6 & ~ & 44.6 & 60.1 & 36.0 & 22.0 & ~ & 62.3 & 20.4 & 40.0\\
MLP-sum-4, Bot.-Up	 & 41.4 & ~ & 67.2 & 41.4 & 54.5 & 32.3 & ~ & 26.1 & 39.3 & 24.3\\
MLP-sum-4, Bot.-Up + aux. loss	 & 46.2 & ~ & 70.9 & 46.2 & 61.5 & 36.0 & ~ & 32.9 & 42.0 & 36.5\\
\Xhline{1\arrayrulewidth}
MLP-cat-4, ResNet	 & 39.1 & ~ & 39.1 & 46.8 & 38.8 & 12.3 & ~ & 61.2 & 11.8 & 34.3\\
MLP-cat-4, ResNet + aux. loss	 & 43.0 & ~ & 43.0 & 52.1 & 43.4 & 13.1 & ~ & 59.0 & 14.1 & 36.4\\
MLP-cat-4, Bot.-Up	 & 46.2 & ~ & 77.3 & 46.2 & 55.0 & 45.0 & ~ & 28.2 & 40.3 & 28.6\\
MLP-cat-4, Bot.-Up + aux. loss	 & 52.5 & ~ & 82.0 & 52.5 & 62.5 & 50.9 & ~ & 38.9 & 45.2 & 41.2\\
\Xhline{1\arrayrulewidth}
MLP-sum-6, ResNet	 & 41.2 & ~ & 41.2 & 55.2 & 33.4 & 16.9 & ~ & 61.6 & 16.6 & 37.5\\
MLP-sum-6, ResNet + aux. loss	 & 41.4 & ~ & 41.4 & 55.9 & 32.9 & 26.6 & ~ & 54.3 & 22.3 & 36.7\\
MLP-sum-6, Bot.-Up	 & 43.8 & ~ & 68.5 & 43.8 & 57.8 & 33.9 & ~ & 23.4 & 41.8 & 24.6\\
MLP-sum-6, Bot.-Up + aux. loss	 & 47.7 & ~ & 73.6 & 47.7 & 62.1 & 35.1 & ~ & 37.4 & 46.1 & 45.1\\
\Xhline{1\arrayrulewidth}
MLP-cat-6, ResNet	 & 40.7 & ~ & 40.7 & 48.2 & 39.5 & 13.7 & ~ & 63.6 & 13.2 & 37.0\\
MLP-cat-6, ResNet + aux. loss	 & 44.5 & ~ & 44.5 & 53.5 & 43.9 & 13.1 & ~ & 61.9 & 13.2 & 39.2\\
MLP-cat-6, Bot.-Up	 & 50.4 & ~ & 80.6 & 50.4 & 59.7 & 48.6 & ~ & 37.2 & 44.0 & 40.4\\
MLP-cat-6, Bot.-Up + aux. loss	 & 55.7 & ~ & 83.8 & 55.7 & 65.6 & 53.6 & ~ & 43.0 & 48.9 & 44.7\\
\Xhline{1\arrayrulewidth}
GRU, ResNet	 & 20.8 & ~ & 20.8 & 24.7 & 20.2 & 12.6 & ~ & 18.1 & 13.1 & 18.4\\
GRU, ResNet + aux. loss	 & 31.0 & ~ & 31.0 & 39.8 & 28.5 & 13.0 & ~ & 29.2 & 13.3 & 26.5\\
GRU, Bot.-Up	 & 43.8 & ~ & 77.4 & 43.8 & 56.8 & 40.0 & ~ & 20.8 & 37.0 & 20.7\\
GRU, Bot.-Up + aux. loss	 & 50.6 & ~ & 81.6 & 50.6 & 63.1 & 49.2 & ~ & 26.0 & 41.9 & 27.9\\
\Xhline{1\arrayrulewidth}
GRU-shared, ResNet	 & 43.4 & ~ & 43.4 & 55.4 & 40.8 & 14.8 & ~ & 65.8 & 14.5 & 36.8\\
GRU-shared, ResNet + aux. loss	 & 48.2 & ~ & 48.2 & 60.5 & 45.5 & 22.8 & ~ & 65.8 & 19.9 & 41.3\\
GRU-shared, Bot.-Up	 & 46.7 & ~ & 77.1 & 46.7 & 59.6 & 44.1 & ~ & 25.9 & 38.3 & 26.9\\
GRU-shared, Bot.-Up + aux. loss	 & 52.7 & ~ & 82.5 & 52.7 & 67.3 & 49.9 & ~ & 34.2 & 42.0 & 40.3\\
\Xhline{1\arrayrulewidth}
VQA-like, ResNet	 & 36.7 & ~ & 36.7 & 52.2 & 33.1 & 20.7 & ~ & 54.0 & 17.3 & 26.3\\
VQA-like, ResNet + aux. loss	 & 39.7 & ~ & 39.7 & 57.1 & 36.2 & 21.9 & ~ & 55.0 & 20.6 & 27.7\\
VQA-like, Bot.-Up	 & 37.9 & ~ & 59.9 & 37.9 & 54.0 & 33.4 & ~ & 20.2 & 28.1 & 20.4\\
VQA-like, Bot.-Up + aux. loss	 & 41.0 & ~ & 62.4 & 41.0 & 57.8 & 35.8 & ~ & 24.6 & 31.0 & 25.0\\
\Xhline{1\arrayrulewidth}
RN without panel IDs, ResNet	 & 32.6 & ~ & 32.6 & 35.8 & 27.3 & 14.0 & ~ & 56.2 & 13.7 & 36.6\\
RN without panel IDs, ResNet + aux. loss	 & 35.0 & ~ & 35.0 & 39.5 & 27.1 & 16.7 & ~ & 57.9 & 15.3 & 40.2\\
RN without panel IDs, Bot.-Up	 & 35.8 & ~ & 46.4 & 35.8 & 40.6 & 28.0 & ~ & 15.1 & 40.9 & 14.0\\
RN without panel IDs, Bot.-Up + aux. loss	 & 38.0 & ~ & 47.5 & 38.0 & 43.6 & 28.7 & ~ & 18.3 & 44.0 & 15.4\\
\Xhline{1\arrayrulewidth}
RN, ResNet	 & 51.2 & ~ & 51.2 & 64.3 & 43.0 & 20.7 & ~ & 78.8 & 22.1 & 49.3\\
RN, ResNet + aux. loss	 & 55.8 & ~ & 55.8 & 69.8 & 44.3 & 34.2 & ~ & 78.2 & 28.3 & 55.4\\
RN, Bot.-Up	 & 55.4 & ~ & 83.0 & 55.4 & 68.3 & 46.2 & ~ & 31.3 & 54.0 & 33.6\\
RN, Bot.-Up + aux. loss	 & 61.3 & ~ & 88.2 & 61.3 & 76.8 & 48.4 & ~ & 41.9 & 60.3 & 45.8\\
\Xhline{1\arrayrulewidth}
\end{tabularx}
\normalsize
\normalsize
\caption{Evaluation of all models in the neutral setting.}
\label{tableNeutral}
\end{table*}

\begin{table*}[t]
\small
\renewcommand{\tabcolsep}{0.15em}
\renewcommand{\arraystretch}{1.18}
\centering
\begin{tabularx}{\linewidth}{Xc c cccccc}
\Xhline{1\arrayrulewidth}
                                          &   Neutral & ~~ & \multicolumn{6}{c}{Generalization settings} \vspace{-1pt}\\
                                          &   setting & ~~ & \textit{interpolation} & \textit{extrapolation} & \textit{att.held} & \textit{att.rel.held} & \textit{obj.rel.held} & \textit{obj.held}\\
\Xhline{1\arrayrulewidth}
Human evaluation & 77.8 & ~ & -- & -- & -- & -- & -- & --\\
Lower bound: RN (last row) with shuffled inputs & 12.5 & ~ & 12.5 & 12.5 & 12.5 & 12.5 & 12.5 & 12.5\\
\Xhline{1\arrayrulewidth}
MLP-sum-2, ResNet	 & 35.1 & ~ & 28.8 & 30.3 & 29.3 & 31.6 & 31.9 & 32.1\\
MLP-sum-2, ResNet + aux. loss	 & 39.1 & ~ & 33.7 & 29.7 & 29.2 & 35.0 & 34.7 & 36.5\\
MLP-sum-2, Bot.-Up	 & 36.4 & ~ & 30.3 & 28.0 & 29.5 & 31.0 & 31.3 & 35.4\\
MLP-sum-2, Bot.-Up + aux. loss	 & 40.4 & ~ & 31.8 & 29.0 & 29.5 & 33.4 & 34.7 & 38.1\\
\Xhline{1\arrayrulewidth}
MLP-cat-2, ResNet	 & 34.7 & ~ & 27.6 & 27.6 & 26.3 & 29.1 & 30.5 & 31.2\\
MLP-cat-2, ResNet + aux. loss	 & 39.2 & ~ & 30.8 & 29.8 & 27.1 & 34.7 & 35.0 & 35.4\\
MLP-cat-2, Bot.-Up	 & 37.6 & ~ & 30.3 & 29.2 & 29.1 & 30.3 & 36.1 & 35.8\\
MLP-cat-2, Bot.-Up + aux. loss	 & 41.7 & ~ & 31.7 & 30.8 & 29.6 & 33.4 & 38.7 & 39.1\\
\Xhline{1\arrayrulewidth}
MLP-sum-4, ResNet	 & 37.6 & ~ & 32.3 & 31.2 & 13.5 & 32.1 & 33.8 & 34.0\\
MLP-sum-4, ResNet + aux. loss	 & 44.6 & ~ & 35.2 & 35.4 & 29.8 & 37.0 & 41.3 & 37.8\\
MLP-sum-4, Bot.-Up	 & 41.4 & ~ & 32.7 & 29.9 & 33.1 & 34.5 & 34.2 & 37.5\\
MLP-sum-4, Bot.-Up + aux. loss	 & 46.2 & ~ & 36.1 & 33.0 & 31.5 & 36.0 & 38.7 & 41.3\\
\Xhline{1\arrayrulewidth}
MLP-cat-4, ResNet	 & 39.1 & ~ & 32.0 & 33.5 & 31.3 & 34.9 & 34.3 & 34.4\\
MLP-cat-4, ResNet + aux. loss	 & 43.0 & ~ & 38.1 & 37.1 & 31.9 & 39.7 & 40.5 & 39.4\\
MLP-cat-4, Bot.-Up	 & 46.2 & ~ & 38.1 & 36.4 & 33.1 & 37.0 & 43.4 & 42.4\\
MLP-cat-4, Bot.-Up + aux. loss	 & 52.5 & ~ & 40.9 & 40.4 & 34.3 & 40.5 & 47.0 & 48.1\\
\Xhline{1\arrayrulewidth}
MLP-sum-6, ResNet	 & 41.2 & ~ & 33.4 & 32.9 & 18.7 & 32.0 & 35.0 & 37.1\\
MLP-sum-6, ResNet + aux. loss	 & 41.4 & ~ & 36.6 & 37.4 & 32.0 & 35.9 & 41.7 & 39.7\\
MLP-sum-6, Bot.-Up	 & 43.8 & ~ & 32.5 & 30.0 & 32.5 & 33.9 & 36.3 & 41.0\\
MLP-sum-6, Bot.-Up + aux. loss	 & 47.7 & ~ & 37.8 & 35.3 & 33.6 & 36.8 & 41.1 & 42.6\\
\Xhline{1\arrayrulewidth}
MLP-cat-6, ResNet	 & 40.7 & ~ & 35.2 & 33.8 & 30.5 & 36.2 & 34.4 & 35.8\\
MLP-cat-6, ResNet + aux. loss	 & 44.5 & ~ & 38.4 & 40.2 & 33.6 & 38.9 & 43.6 & 40.5\\
MLP-cat-6, Bot.-Up	 & 50.4 & ~ & 40.4 & 39.0 & 36.4 & 39.3 & 46.9 & 45.8\\
MLP-cat-6, Bot.-Up + aux. loss	 & 55.7 & ~ & 43.3 & 43.9 & 35.8 & 42.7 & 54.0 & 51.3\\
\Xhline{1\arrayrulewidth}
GRU, ResNet	 & 20.8 & ~ & 30.0 & 23.9 & 25.4 & 12.6 & 20.8 & 12.7\\
GRU, ResNet + aux. loss	 & 31.0 & ~ & 36.5 & 33.7 & 25.9 & 12.6 & 24.4 & 23.5\\
GRU, Bot.-Up	 & 43.8 & ~ & 36.3 & 32.8 & 31.2 & 23.6 & 34.7 & 43.3\\
GRU, Bot.-Up + aux. loss	 & 50.6 & ~ & 38.4 & 37.5 & 32.2 & 33.4 & 40.1 & 48.0\\
\Xhline{1\arrayrulewidth}
GRU-shared, ResNet	 & 43.4 & ~ & 34.2 & 33.0 & 34.4 & 38.4 & 37.0 & 41.2\\
GRU-shared, ResNet + aux. loss	 & 48.2 & ~ & 38.1 & 37.3 & 35.2 & 40.0 & 42.2 & 44.8\\
GRU-shared, Bot.-Up	 & 46.7 & ~ & 36.8 & 36.4 & 35.4 & 37.4 & 41.1 & 44.5\\
GRU-shared, Bot.-Up + aux. loss	 & 52.7 & ~ & 39.7 & 38.8 & 36.4 & 41.5 & 47.6 & 48.8\\
\Xhline{1\arrayrulewidth}
VQA-like, ResNet	 & 36.7 & ~ & 28.1 & 27.7 & 29.4 & 34.0 & 33.2 & 35.0\\
VQA-like, ResNet + aux. loss	 & 39.7 & ~ & 30.2 & 29.7 & 31.5 & 35.6 & 36.4 & 37.8\\
VQA-like, Bot.-Up	 & 37.9 & ~ & 28.2 & 28.1 & 31.8 & 32.7 & 33.8 & 35.5\\
VQA-like, Bot.-Up + aux. loss	 & 41.0 & ~ & 30.6 & 30.9 & 31.0 & 35.1 & 36.2 & 38.1\\
\Xhline{1\arrayrulewidth}
RN without panel IDs, ResNet	 & 32.6 & ~ & 27.7 & 28.1 & 30.4 & 32.0 & 30.9 & 30.6\\
RN without panel IDs, ResNet + aux. loss	 & 35.0 & ~ & 29.7 & 27.3 & 31.2 & 32.6 & 32.9 & 31.8\\
RN without panel IDs, Bot.-Up	 & 35.8 & ~ & 31.1 & 29.1 & 32.6 & 32.7 & 32.2 & 32.5\\
RN without panel IDs, Bot.-Up + aux. loss	 & 38.0 & ~ & 31.7 & 31.6 & 31.7 & 34.2 & 34.9 & 33.2\\
\Xhline{1\arrayrulewidth}
RN, ResNet	 & 51.2 & ~ & 39.8 & 39.0 & 42.1 & 43.4 & 48.9 & 47.5\\
RN, ResNet + aux. loss	 & 55.8 & ~ & 42.4 & 42.3 & 40.9 & 47.3 & 52.8 & 51.0\\
RN, Bot.-Up	 & 55.4 & ~ & 43.4 & 39.7 & 43.6 & 42.2 & 51.9 & 50.6\\
RN, Bot.-Up + aux. loss	 & 61.3 & ~ & 47.4 & 45.2 & 44.1 & 44.9 & 58.5 & 55.2\\
\Xhline{1\arrayrulewidth}
\end{tabularx}
\normalsize
\normalsize
\caption{Overall accuracy of all models trained/evaluated on splits requiring varying levels of generalization.}
\label{tableGeneralization}
\end{table*}

\section{Dataset examples}

We provide in \fig\ref{fig:detresults} a random selection of instances from our dataset.

\clearpage

\begin{figure*}[ht]
 \centering
 \subfigure[]{ \includegraphics[height=2.3in,width=2.2in]{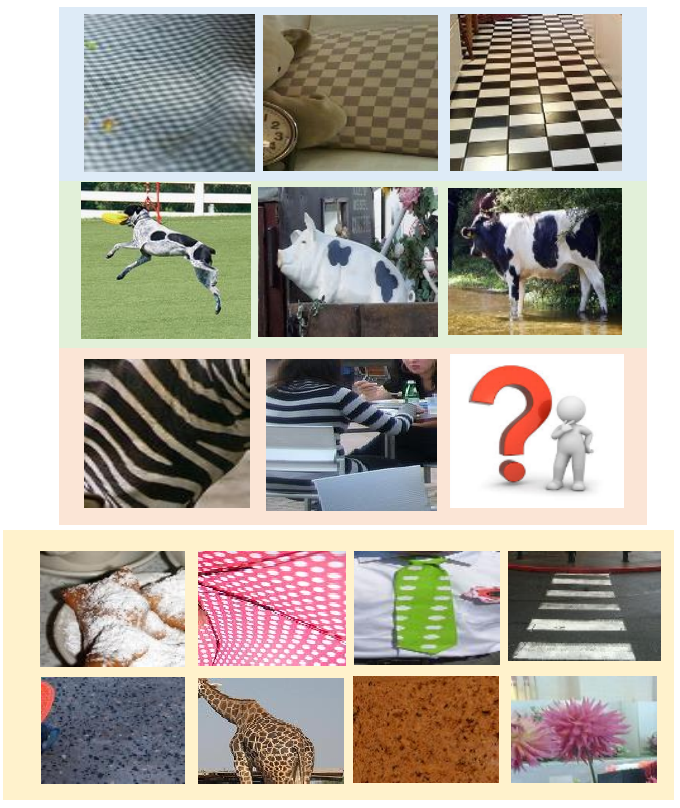}} 
 \subfigure[]{ \includegraphics[height=2.3in,width=2.2in]{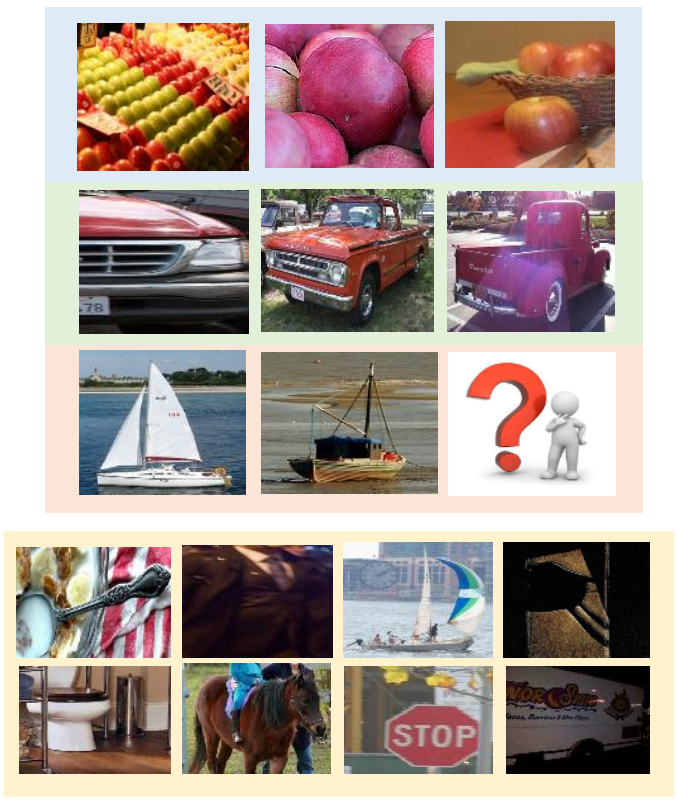}} 
 \subfigure[]{ \includegraphics[height=2.3in,width=2.2in]{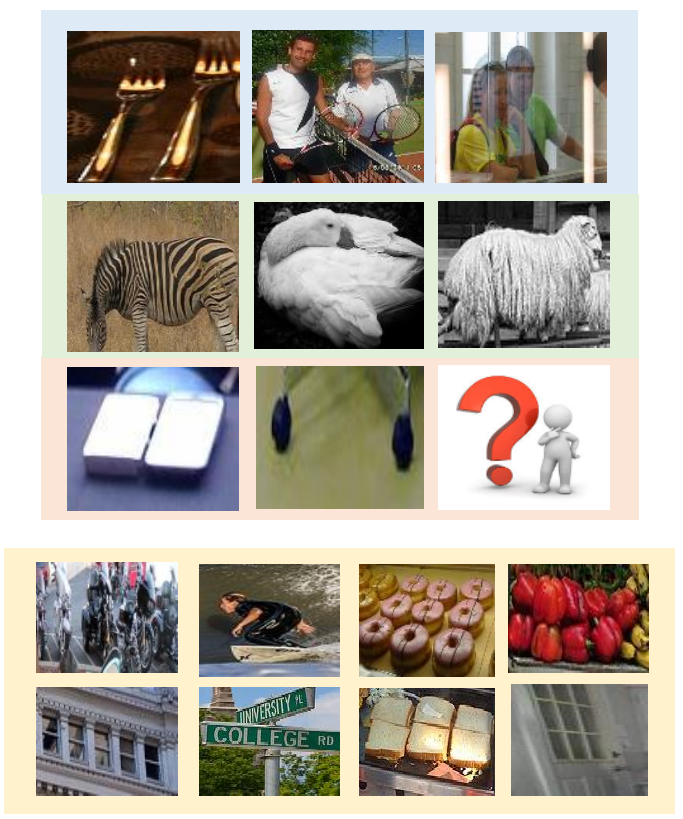}} \\
 \subfigure[]{ \includegraphics[height=2.3in,width=2.2in]{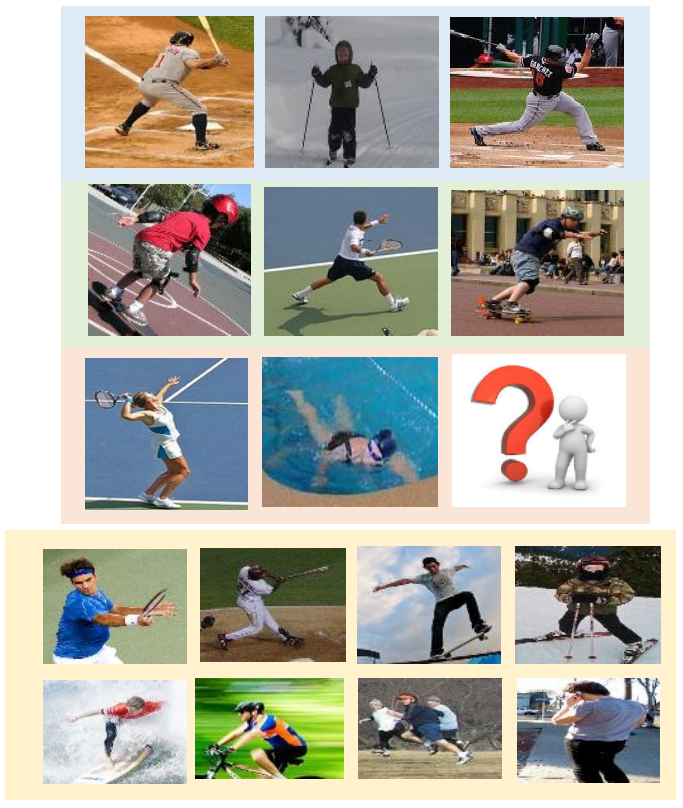}} 
 \subfigure[]{ \includegraphics[height=2.3in,width=2.2in]{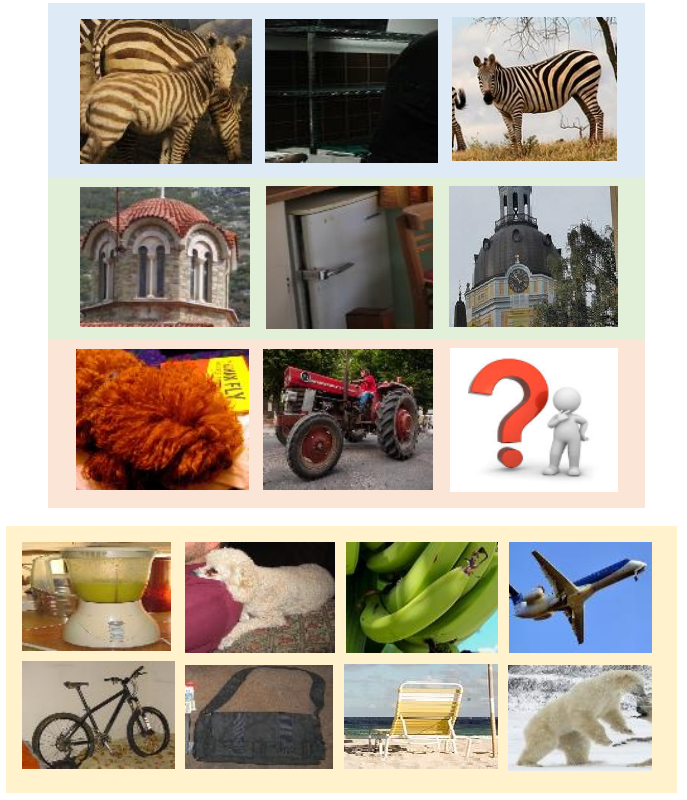}} 
 \subfigure[]{ \includegraphics[height=2.3in,width=2.2in]{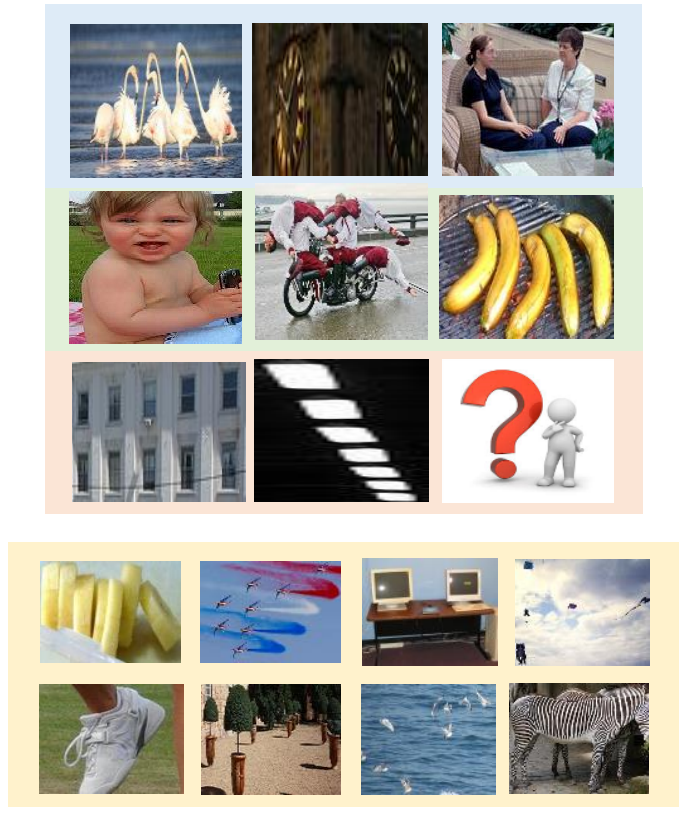}} \\
 \subfigure[]{ \includegraphics[height=2.3in,width=2.2in]{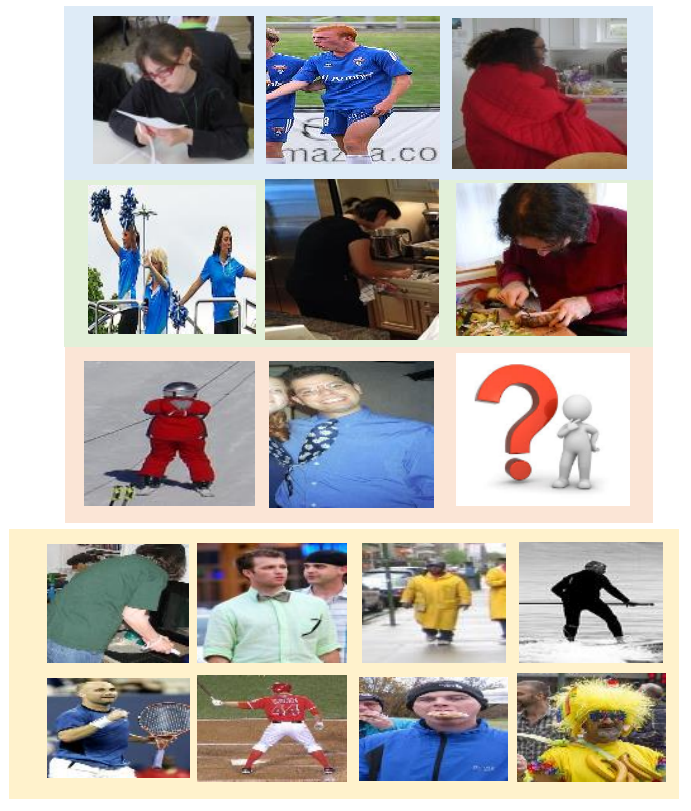}} 
 \subfigure[]{ \includegraphics[height=2.3in,width=2.2in]{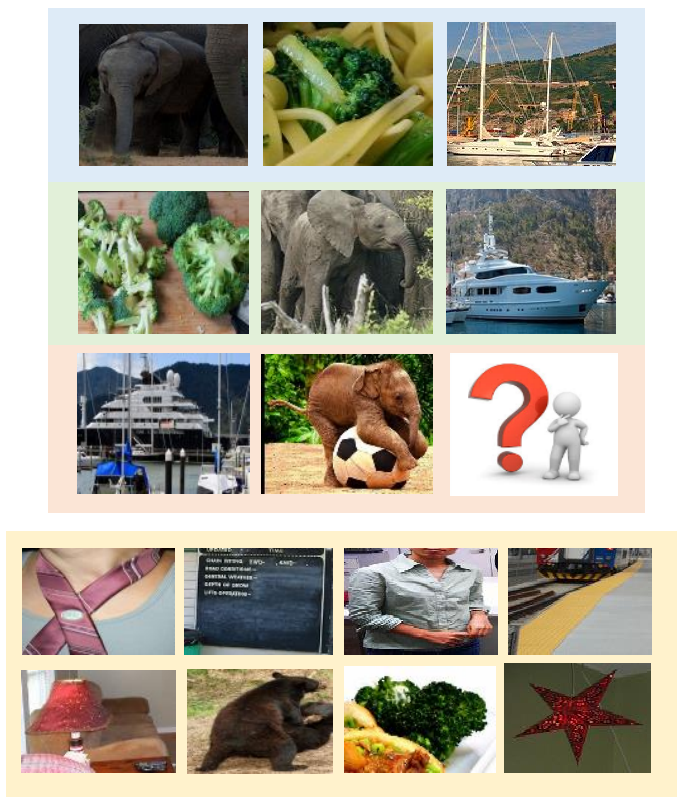}} 
 \subfigure[]{ \includegraphics[height=2.3in,width=2.2in]{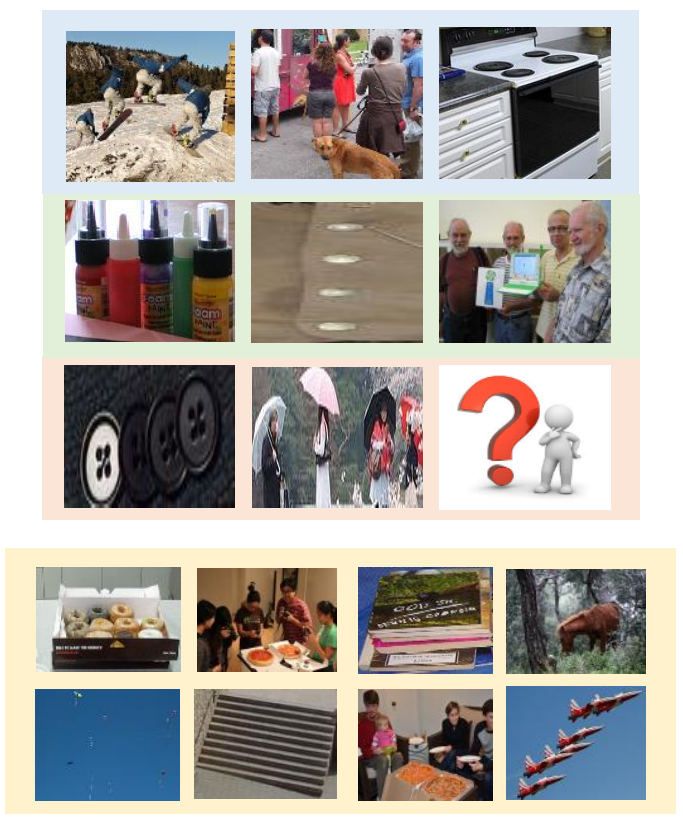}} 
 \end{figure*}
 \addtocounter{figure}{-1}
 \begin{figure*}[t]
 \centering
 \setcounter{figure}{10}
 \subfigure[]{ \includegraphics[height=2.3in,width=2.2in]{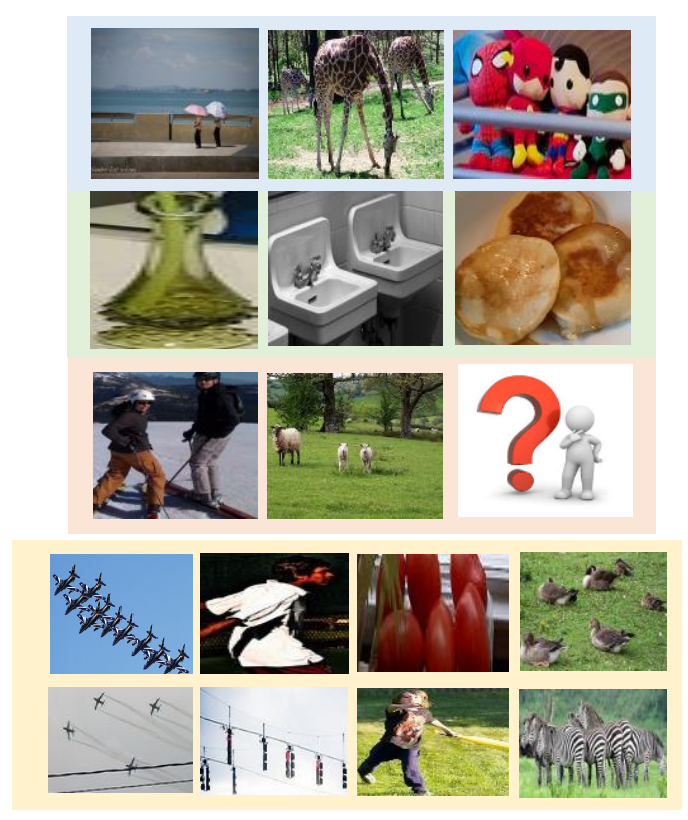}} 
 \subfigure[]{ \includegraphics[height=2.3in,width=2.2in]{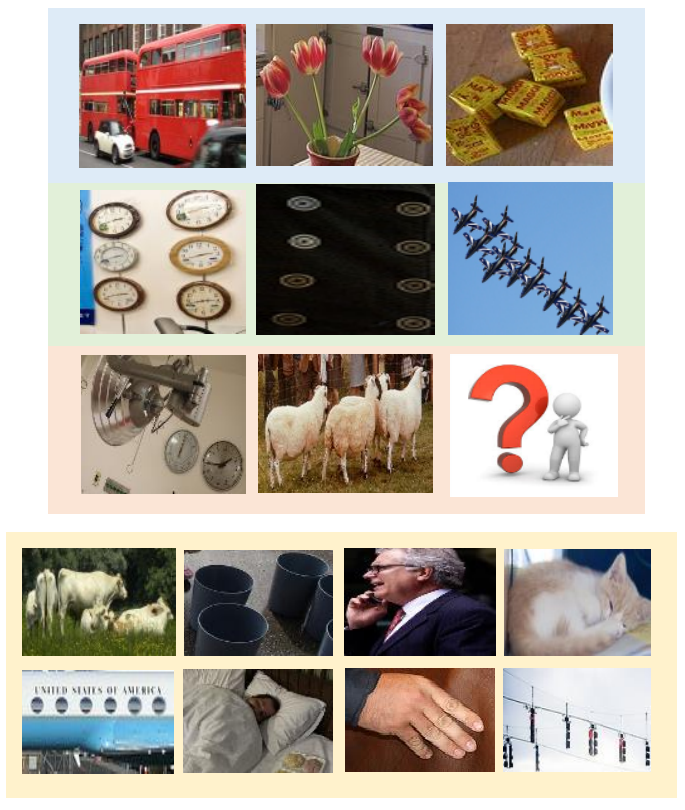}} 
 \subfigure[]{ \includegraphics[height=2.3in,width=2.2in]{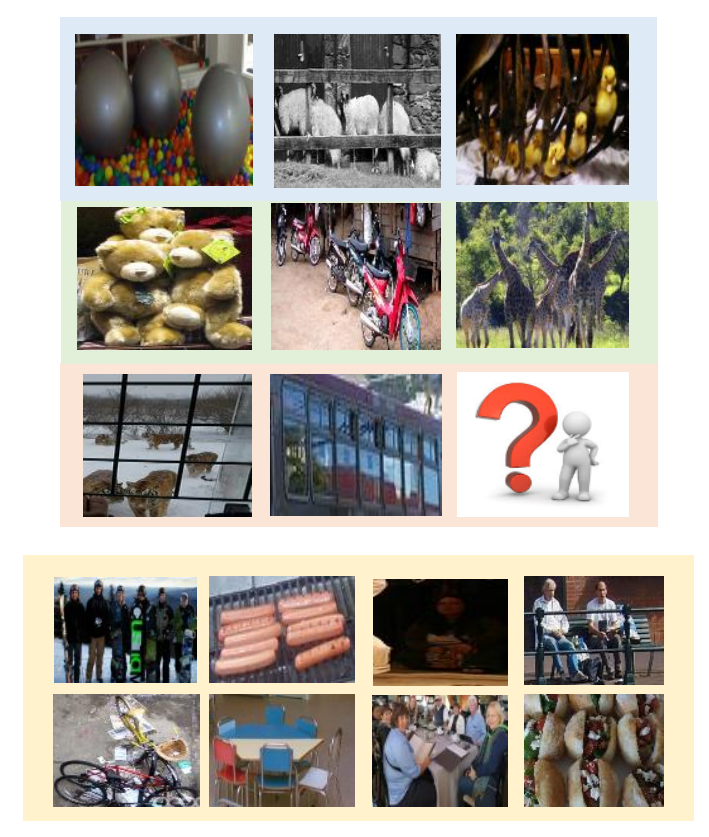}} 
 \vspace{0.5em}
 \caption{Additional examples from our dataset. The four rows respecitvely depict the relations \textit{And}, \textit{Or}, \textit{Union}, and \textit{Progression}. The correct answers are 4,3,6,1,2,1,4,7,2,5,5,8, referring to the candidate answers as 1--8 left to right, first then second row.} \label{fig:detresults}
\end{figure*}

\end{document}